\documentclass{article}
\usepackage{amsmath}
\usepackage{subfigure}

     \PassOptionsToPackage{numbers, compress}{natbib}

     \usepackage[final]{neurips_2019}

\usepackage[utf8]{inputenc} 
\usepackage[T1]{fontenc}    
\usepackage{hyperref}       
\usepackage{url}            
\usepackage{booktabs}       
\usepackage{amsfonts}       
\usepackage{nicefrac}       
\usepackage{microtype}      

\title{Keep It Simple: Graph Autoencoders Without Graph~Convolutional~Networks}
\author{%
  Guillaume Salha\\
Deezer Research and Development\\
  LIX, École Polytechnique\\
   \And
     Romain Hennequin\\
  Deezer Research and Development\\
     \AND
     Michalis Vazirgiannis\\
  LIX, École Polytechnique \& AUEB\\
}

\begin{document}

\maketitle

\begin{abstract}
Graph autoencoders (AE) and variational autoencoders (VAE) recently emerged as powerful node embedding methods, with promising performances on challenging tasks such as link prediction and node clustering. Graph AE, VAE and most of their extensions rely on graph convolutional networks (GCN) to learn vector space representations of nodes. In this paper, we propose to replace the GCN encoder by a simple linear model w.r.t. the adjacency matrix of the graph. For the two aforementioned tasks, we empirically show that this approach consistently reaches competitive performances w.r.t. GCN-based models for numerous real-world graphs, including the widely used Cora, Citeseer and Pubmed citation networks that became the \textit{de facto} benchmark datasets for evaluating graph AE and VAE. This result questions the relevance of repeatedly using these three datasets to compare complex graph AE and VAE models. It also emphasizes the effectiveness of simple node encoding schemes for many real-world applications.
\end{abstract}

\section{Introduction}

Graphs have become ubiquitous, due to the proliferation of data representing relationships or interactions among entities \cite{hamilton2017representation, wu2019comprehensive}. Extracting relevant information from these entities, called the \textit{nodes} of the graph, is crucial to effectively tackle numerous machine learning tasks, such as link prediction or node clustering. While traditional approaches mainly focused on hand-engineered features \cite{bhagat2011node, liben2007link}, significant improvements were recently achieved by methods aiming at directly \textit{learning} node representations that summarize the graph structure (see \cite{hamilton2017representation} for a review). In a nutshell, these \textit{representation learning} methods aim at embedding nodes as vectors in a low-dimensional vector space in which nodes with structural proximity in the graph should be close, e.g. by leveraging random walk strategies \cite{ grover2016node2vec, perozzi2014deepwalk}, matrix factorization \cite{cao2015grarep, ou2016asymmetric} or graph neural networks \cite{hamilton2017inductive, kipf2016-1}.

In particular, \textit{graph autoencoders} (AE) \cite{ kipf2016-2, tian2014learning, wang2016structural} and \textit{variational autoencoders} (VAE) \cite{kipf2016-2} recently emerged as powerful node embedding methods. Based on encoding-decoding schemes, i.e. on the design of low dimensional vector space representations of nodes (\textit{encoding}) from which reconstructing the original graph structure (\textit{decoding}) is possible, graph AE and VAE models have been successfully applied to confront several challenging learning tasks, with competitive results w.r.t. popular baselines such as \cite{grover2016node2vec, perozzi2014deepwalk,tang2015line}. These tasks include link prediction \cite{grover2019graphite, kipf2016-2, pan2018arga, tran2018multi, salha2019-1, salha2019-2}, node clustering \cite{pan2018arga, salha2019-1, wang2017mgae}, matrix completion for inference and recommendation \cite{berg2018matrixcomp, do2019matrix} and molecular graph generation \cite{molecule3, molecule1,molecule2,simonovsky2018graphvae}. Existing models usually rely on graph neural networks (GNN) to encode nodes into the embedding ; more precisely, most of them implement \textit{graph convolutional networks} (GCN) encoders \cite{do2019matrix, grover2019graphite,  huang2019rwr, kipf2016-2, lerique2019joint, pan2018arga, salha2019-1, salha2019-2}, a model originally introduced in \cite{kipf2016-1}.

In this paper, we analyse the empirical benefit of including GCN encoders in graph AE and VAE w.r.t. more simple heuristics. After reviewing key concepts in Section 2, we introduce simpler versions of graph AE and VAE in Section 3, replacing GCNs by a straightforward linear model w.r.t. the adjacency matrix of the graph, involving a unique weight matrix. In Section 4, we show that these models are empirically competitive\footnote{We publicly release the code of these experiments at: \href{https://github.com/deezer/linear_graph_autoencoders}{https://github.com/deezer/linear\_graph\_autoencoders}} w.r.t. GCN-based graph AE and VAE on link prediction and node clustering tasks, for numerous real-world datasets. We discuss these results, aiming at providing insights on settings where GCN encoders should bring (or not) empirical benefits.


\section{Preliminaries on Graph (Variational) Autoencoders}

Throughout this paper, we consider an undirected graph $\mathcal{G} = (\mathcal{V},\mathcal{E})$ with $|\mathcal{V}| = n$ nodes and $|\mathcal{E}| = m$ edges. We denote by $A$ the  adjacency matrix of $\mathcal{G}$, that is either binary or weighted. 

\paragraph{Graph Autoencoders} Graph autoencoders (AE) \cite{ kipf2016-2, tian2014learning, wang2016structural} are a family of models aiming at mapping (\textit{encoding}) each node $i \in \mathcal{V}$ to a vector $z_i \in \mathbb{R}^d$, with $d \ll n$, from which reconstructing (\textit{decoding}) the graph should be possible. Intuitively, if, starting from the node embedding, the model is also able to reconstruct an adjacency matrix $\hat{A}$ close to the true one, then the $z_i$ vectors should capture some important characteristics of the graph structure. More precisely, the $n \times d$ matrix $Z$ of all $z_i$ vectors is usually the output of a graph neural network (GNN) \cite{bruna2013spectral, defferrard2016, kipf2016-1} processing $A$. To reconstruct the graph, we stack an inner product decoder to this GNN, as in most models \cite{kipf2016-2}. We have $\hat{A}_{ij} = \sigma (z^T_i z_j)$ for all node pairs $(i,j)$, with $\sigma(\cdot)$ denoting the sigmoid function: $\sigma(x) = 1/(1 + e^{-x})$. Therefore, the larger the inner product $z^T_i z_j$ in the embedding, the more likely nodes $i$ and $j$ are connected in $\mathcal{G}$ according to the AE. To sum up, we have $\hat{A} = \sigma(ZZ^T)$ with $Z = \text{GNN(A)}$. Weights of the GNN are trained by gradient descent \cite{goodfellow2016deep} to minimize a \textit{reconstruction loss} capturing the similarity of $A$ and $\hat{A}$, usually formulated as a weighted cross entropy loss \cite{kipf2016-2}.

\paragraph{Graph Variational Autoencoders} \cite{kipf2016-2} also extended the \textit{variational autoencoder} (VAE) framework \cite{kingma2013vae} to graph structures. Authors designed a probabilistic model involving latent variables $z_i$ of length $d \ll n$ for each node $i \in \mathcal{V}$, interpreted as node representations in an embedding space. The inference model, i.e. the \textit{encoding} part of the VAE, is defined as:
$$q(Z|A) = \prod_{i=1}^n q(z_i|A) \hspace{10pt} \text{where} \hspace{10pt} q(z_i|A) = \mathcal{N}(z_i|\mu_i, \text{diag}(\sigma_i^2)).$$ Gaussian parameters are learned from two GNNs, i.e. $\mu = \text{GNN}_{\mu}(A)$, with $\mu$ the matrix stacking up mean vectors $\mu_i$ ;  likewise, $\log \sigma = \text{GNN}_{\sigma}(A)$. Latent vectors $z_i$ are samples drawn from this distribution. From these vectors, a generative model aims at reconstructing (\textit{decoding}) $A$, leveraging inner products: $p(A|Z) = \prod_{i=1}^n \prod_{j=1}^n p(A_{ij}|z_i, z_j)$, where  $p(A_{ij} = 1|z_i, z_j) = \sigma(z_i^Tz_j)$. During training, GNN weights are tuned by maximizing a tractable variational lower bound (ELBO) of the model's likelihood (see \cite{kipf2016-2} for details)
by gradient descent, with a Gaussian prior on the distribution of latent vectors, and using the \textit{reparameterization trick} from \cite{kingma2013vae}.

\paragraph{Graph Convolutional Networks} While the term \textit{GNN encoder} is generic, a majority of successful applications and extensions of graph AE and VAE \cite{do2019matrix, grover2019graphite,  huang2019rwr, kipf2016-2, lerique2019joint, pan2018arga, salha2019-1, salha2019-2} actually relied on \textit{graph convolutional networks (GCN)} \cite{kipf2016-1} to encode nodes, including the seminal models from \cite{kipf2016-2}. 
In a GCN with $L$ layers ($L \geq 2$), with input layer $H^{(0)} = I_n$ and output layer $H^{(L)}$ ($H^{(L)} = Z$ for AE, and $\mu$ or $\log \sigma$ for VAE), embedding vectors are iteratively updated, as follows:
$$H^{(l)} = \text{ReLU} (\tilde{A} H^{(l-1)} W^{(l-1)}), \hspace{3pt} \text{for } l \in \{1,...L-1\} \hspace{10pt} \text{and} \hspace{10pt} H^{(L)} = \tilde{A} H^{(L-1)} W^{(L-1)},$$
where $\tilde{A} = D^{-1/2}(A + I_n) D^{-1/2}$. $D$ is the diagonal degree matrix of $A + I_n$, and $\tilde{A}$ is therefore its symmetric normalization. At each layer, each node averages representations from its neighbors (that, from layer 2, have aggregated representations from their own neighbors), with a ReLU activation: $\text{ReLU}(x) = \max(x,0)$. Matrices $W^{(0)},...,W^{(L-1)}$, whose dimensions can vary, are weight matrices to tune. GCN became a popular encoding scheme, thanks to its relative simplicity w.r.t. \cite{bruna2013spectral,defferrard2016} and thanks to the linear complexity w.r.t. $m$ of evaluating each layer \cite{kipf2016-1}. Last, GCN models can also leverage node-level features, summarized in an $n \times f$ matrix $X$, in addition to the graph structure. In such setting, the input layer becomes $H^{(0)} = X$ instead of the identity matrix $I_n$.

\section{Simplifying the Encoding Scheme}

\paragraph{Linear Graph AE} In this section, we propose to replace the GCN encoder by a simple linear model w.r.t. the normalized adjacency matrix of the graph. In the AE framework, we have:
$$ Z = \tilde{A}  W \hspace{10pt} \text{then} \hspace{10pt}\hat{A} = \sigma(ZZ^T).$$
Embedding vectors are obtained by multiplying the $n \times n$ normalized adjacency matrix $\tilde{A}$ by a single $n \times d$ weight matrix $W$, tuned by gradient descent in a similar fashion w.r.t. standard AE. This encoder is a straightforward linear mapping. Contrary to standard GCN encoders (as $L\geq 2$), nodes only aggregate information from their one-step neighbors. If data include node-level features $X$, the encoding step becomes $Z = \tilde{A} X W$, and $W$ is of dimension $f \times d$.

\paragraph{Linear Graph VAE} We adopt the same approach to replace the encoder of graph VAE by:
$$ \mu = \tilde{A} W_{\mu} \hspace{3pt} \text{and} \hspace{3pt} \log \sigma = \tilde{A} W_{\sigma} \hspace{10pt} \text{then} \hspace{10pt}\forall i \in \mathcal{V}, z_i \sim \mathcal{N}(\mu_i, \text{diag}(\sigma_i^2)),$$
with similar decoder w.r.t. standard graph VAE. We refer to this simpler model as \textit{linear graph VAE}, and optimize the standard ELBO bound \cite{kipf2016-2} w.r.t. weight matrices $W_{\mu}$ and $W_{\sigma}$ by gradient descent. If data include node-level features $X$, we compute $\mu = \tilde{A} X W_{\mu}$ and $\log \sigma = \tilde{A} X W_{\sigma}$.

\section{Empirical Analysis and Discussion}

\begin{table}[t]
  \centering
    \caption{Link prediction on Cora, Citeseer and Pubmed. Performances of linear graph AE/VAE are \underline{underlined} when reaching competitive results  w.r.t. GCN-based models from \cite{kipf2016-2} ($\pm$ 1 st. dev.).}
\begin{tiny}
  \begin{tabular}{c|cc|cc|cc}
    \toprule
    & \multicolumn{2}{c}{\textbf{Cora}} & \multicolumn{2}{c}{\textbf{Citeseer}} & \multicolumn{2}{c}{\textbf{Pubmed}} \\
     \textbf{Model} & \multicolumn{2}{c}{(n = 2 708, m = 5 429)} & \multicolumn{2}{c}{(n = 3 327, m = 4 732)} & \multicolumn{2}{c}{(n = 19 717, m = 44 338)} \\
     \cmidrule{2-7}
      & \tiny \textbf{AUC (in \%)} & \tiny \textbf{AP (in \%)} & \tiny \textbf{AUC (in \%)} & \tiny \textbf{AP (in \%)} & \tiny \textbf{AUC (in \%)} & \tiny \textbf{AP (in \%)}\\
    \midrule
    \midrule
     Linear AE (ours) & \underline{83.19} $\pm$ \underline{1.13} & \underline{87.57} $\pm$ \underline{0.95} & \underline{77.06} $\pm$ \underline{1.81} & \underline{83.05} $\pm$ \underline{1.25} & 81.85 $\pm$ 0.32 & \underline{87.54} $\pm$ \underline{0.28} \\
     2-layer GCN AE& 84.79 $\pm$ 1.10 & 88.45 $\pm$ 0.82 & 78.25 $\pm$ 1.69 & 83.79 $\pm$ 1.24 & 82.51 $\pm$ 0.64 & 87.42 $\pm$ 0.38 \\
     3-layer GCN AE & 84.61 $\pm$ 1.22 & 87.65 $\pm$ 1.11 & 78.62 $\pm$ 1.74 & 82.81 $\pm$ 1.43 & 83.37 $\pm$ 0.98 & 87.62 $\pm$ 0.68 \\
    \midrule
    Linear VAE (ours)& \underline{84.70} $\pm$ \underline{1.24} & \underline{88.24} $\pm$ \underline{1.02} & \underline{78.87} $\pm$ \underline{1.34} & \underline{83.34} $\pm$ \underline{0.99} & \underline{84.03} $\pm$ \underline{0.28} & \underline{87.98} $\pm$ \underline{0.25} \\
     2-layer GCN VAE & 84.19 $\pm$ 1.07 & 87.68 $\pm$ 0.93 & 78.08 $\pm$ 1.40 & 83.31 $\pm$ 1.31 & 82.63 $\pm$ 0.45 & 87.45 $\pm$ 0.34 \\
     3-layer GCN VAE & 84.48 $\pm$ 1.42 & 87.61 $\pm$ 1.08 & 79.27 $\pm$ 1.78 & 83.73 $\pm$ 1.13 & 84.07 $\pm$ 0.47 & 88.18 $\pm$ 0.31 \\
    \midrule
    \midrule
       & \multicolumn{2}{c}{\textbf{Cora, with features}} & \multicolumn{2}{c}{\textbf{Citeseer, with features}} & \multicolumn{2}{c}{\textbf{Pubmed, with features}} \\
            \textbf{Model}  & \multicolumn{2}{c}{(n = 2 708, m = 5 429, f = 1 433)} & \multicolumn{2}{c}{(n = 3 327, m = 4 732, f = 3 703)} & \multicolumn{2}{c}{(n = 19 717, m = 44 338, f = 500)} \\
     \cmidrule{2-7}
      & \tiny \textbf{AUC (in \%)} & \tiny \textbf{AP (in \%)} & \tiny \textbf{AUC (in \%)} & \tiny \textbf{AP (in \%)} & \tiny \textbf{AUC (in \%)} & \tiny \textbf{AP (in \%)}\\
    \midrule
    \midrule
    Linear AE (ours) & \underline{92.05} $\pm$ \underline{0.93} & \underline{93.32} $\pm$ \underline{0.86} & \underline{91.50} $\pm$ \underline{1.17} & \underline{92.99} $\pm$ \underline{0.97} & \underline{95.88} $\pm$ \underline{0.20} & \underline{95.89} $\pm$ \underline{0.17} \\
     2-layer GCN AE & 91.27 $\pm$ 0.78 & 92.47 $\pm$ 0.71 & 89.76 $\pm$ 1.39 & 90.32 $\pm$ 1.62 & 96.28 $\pm$ 0.36 & 96.29 $\pm$ 0.25 \\
     3-layer GCN AE & 89.16 $\pm$ 1.18 & 90.98 $\pm$ 1.01 & 87.31 $\pm$ 1.74 & 89.60 $\pm$ 1.52 & 94.82 $\pm$ 0.41 & 95.42 $\pm$ 0.26 \\
    \midrule
    Linear VAE (ours)& \underline{92.55} $\pm$ \underline{0.97} & \underline{93.68} $\pm$ \underline{0.68} & \underline{91.60} $\pm$ \underline{0.90} & \underline{93.08} $\pm$ \underline{0.77} & \underline{95.91} $\pm$ \underline{0.13} & \underline{95.80} $\pm$ \underline{0.17} \\
     2-layer GCN VAE& 91.64 $\pm$ 0.92 & 92.66 $\pm$ 0.91 & 90.72 $\pm$ 1.01 & 92.05 $\pm$ 0.97 & 94.66 $\pm$ 0.51 & 94.84 $\pm$ 0.42 \\
     3-layer GCN VAE & 90.53 $\pm$ 0.94 & 91.71 $\pm$ 0.88 & 88.63 $\pm$ 0.95 & 90.20 $\pm$ 0.81 & 92.78 $\pm$ 1.02 & 93.33 $\pm$ 0.91 \\
    \bottomrule
  \end{tabular}

  \end{tiny}
\end{table} 
  \vspace{-0.2cm}

To compare linear graph AE and VAE to GCN-based models, we first focus on \textit{link prediction}, as in \cite{kipf2016-2} and most subsequent works. We train models on incomplete versions of graphs where $15\%$ of edges were randomly removed. Then, we create validation and test sets from removed edges (resp. from $5\%$ and $10\%$ of edges) and from the same number of randomly sampled pairs of unconnected nodes. We evaluate the model's ability to classify edges from non-edges, using the mean \textit{Area Under the Receiver Operating Characteristic (ROC) Curve} (AUC) and \textit{Average Precision} (AP) scores on test sets, averaged over 100 runs with different random train/validation/test splits.

Table 1 reports results for the Cora, Citeseer and Pubmed citation graphs from \cite{cora}, with and without node features corresponding to bag-of-words vectors. These three graphs were used in the original experiments of \cite{kipf2016-2} and then in the wide majority of recent works \cite{grover2019graphite, huang2019rwr, kipf2016-2, lerique2019joint, pan2018arga, park2019symmetric, salha2019-1, salha2019-2, tran2018multi, wang2017mgae}, becoming the \textit{de facto} benchmark datasets for evaluating graph AE and VAE. For standard AE and VAE, we managed to reproduce performances from \cite{kipf2016-2} ; for all models, we detail hyperparameters in annex. In Table 1, we show that linear models consistently reach competitive performances w.r.t. with 2 and 3-layer GCN encoders, i.e. they are at least as good ($\pm$ 1 standard deviation). In some settings, linear AE/VAE are even slightly better (e.g. $+1.25$ points in AUC for linear VAE on Pubmed with features). We did not report performances of deeper models, due to significant scores deterioration. These results emphasize the effectiveness of the proposed simple encoding scheme, limiting to linear and first-order interactions, on these datasets. In supplementary materials, we report two additional tables, consolidating our results by reaching similar conclusions when replacing inner products by more complex decoders \cite{grover2019graphite,salha2019-2}, and for a \textit{node clustering} task.

In addition to Cora, Citeseer and Pubmed, Table 2 reports experiments on eight other real-world graphs with various characteristics. Hyperparameters details are reported in annex. Overall, the linear graph AE and VAE are competitive in six cases out of nine : for the WebKD \cite{cora} hyperlinks web graph, with and without node features (1703-dim bag of words vectors), the Hamsterster \cite{konect} social network, a larger version of Cora \cite{konect}, and the DBLP \cite{konect} and Arxiv-HepTh \cite{snap} citation networks. Such results confirm the empirical effectiveness of simple node encoding schemes, that might appear as a suitable alternative to complex encoders for many real-world applications.

Nonetheless, GCN-based models are outperforming on the Blogs \cite{konect} graph of hyperlinks between blogs from the 2004 US election (for VAE on link prediction, and for both AE/VAE on node clustering), on the Proteins \cite{konect} network of proteins interactions, and on the Google \cite{konect} hyperlinks network of web pages within Google's sites. These datasets are relatively \textit{dense} ; among dense graphs, the benefit of GCN encoders also increases with the \textit{size} of the graph. Last, the \textit{nature} of the graph seems crucial. In \textit{citation graphs}, if a reference A in an article B cited by some authors is relevant to their work, authors will likely also cite this reference A (creating a first order link); therefore, in such graphs the impact of high-order interactions is empirically limited (even on the quite dense Arxiv-HepTh graph). As a consequence, we conjecture that larger and denser graphs with intrinsic high-order interactions (e.g. some web graphs) should be better suited that the sparse medium-size Cora, Citeseer and Pubmed citation networks, when comparing and evaluating complex encoders.
\begin{table}[t]
  \centering
\begin{tiny}
   \caption{Link prediction on alternative real-world datasets. Performances of linear graph AE/VAE are \underline{underlined} when reaching competitive results  w.r.t. GCN-based models from \cite{kipf2016-2} ($\pm$ 1 st. dev.).}
  \begin{tabular}{c|cc|cc|cc}
    \toprule
   & \multicolumn{2}{c}{\textbf{WebKD}} & \multicolumn{2}{c}{\textbf{WebKD, with features}} & \multicolumn{2}{c}{\textbf{Hamsterster}} \\
       \textbf{Model}   & \multicolumn{2}{c}{(n = 877, m = 1 608)} & \multicolumn{2}{c}{(n = 877, m = 1 608, f = 1 703)} & \multicolumn{2}{c}{(n = 1 858, m = 12 534)} \\
     \cmidrule{2-7}
       & \tiny \textbf{AUC (in \%)} & \tiny \textbf{AP (in \%)} & \tiny \textbf{AUC (in \%)} & \tiny \textbf{AP (in \%)} & \tiny \textbf{AUC (in \%)} & \tiny \textbf{AP (in \%)}\\
    \midrule
    \midrule
    Linear AE (ours) & \underline{77.20} $\pm$ \underline{2.35} & \underline{83.55} $\pm$ \underline{1.81} &\underline{84.15} $\pm$ \underline{1.64} & \underline{87.01} $\pm$ \underline{1.48} & \underline{93.07} $\pm$ \underline{0.67} & \underline{94.20} $\pm$ \underline{0.58} \\
     2-layer GCN AE& 77.88 $\pm$ 2.57 & 84.12 $\pm$ 2.18 & 86.03 $\pm$ 3.97 & 87.97 $\pm$ 2.76 & 92.07 $\pm$ 0.63 & 93.01 $\pm$ 0.69 \\
     3-layer GCN AE& 78.20 $\pm$ 3.69 & 83.13 $\pm$ 2.58 & 81.39 $\pm$ 3.93 & 85.34 $\pm$ 2.92 & 91.40 $\pm$ 0.79 & 92.22 $\pm$ 0.85 \\
    \midrule
    Linear VAE (ours) & \underline{83.50} $\pm$ \underline{1.98} & \underline{86.70 $\pm$ 1.53} & 85.57 $\pm$ 2.18 & \underline{88.08} $\pm$ \underline{1.76} & \underline{91.08} $\pm$ \underline{0.70} & \underline{91.85} $\pm$ \underline{0.64} \\
     2-layer GCN VAE& 82.31 $\pm$ 2.55 & 86.15 $\pm$ 2.03 & 87.87 $\pm$ 2.48 & 88.97 $\pm$ 2.17 & 91.62 $\pm$ 0.60 & 92.43 $\pm$ 0.64 \\
     3-layer GCN VAE& 82.17 $\pm$ 2.70 & 85.35 $\pm$ 2.25 & 89.69 $\pm$ 1.80 & 89.90 $\pm$ 1.58 & 91.06 $\pm$ 0.71 & 91.85 $\pm$ 0.77 \\
    \midrule
    \midrule
    & \multicolumn{2}{c}{\textbf{DBLP}} & \multicolumn{2}{c}{\textbf{Cora-larger}} & \multicolumn{2}{c}{\textbf{Arxiv-HepTh}} \\
        \textbf{Model} & \multicolumn{2}{c}{(n = 12 591, m = 49 743)} & \multicolumn{2}{c}{(n = 23 166, m = 91 500)} & \multicolumn{2}{c}{(n = 27 770, m = 352 807)} \\
     \cmidrule{2-7}
      & \tiny \textbf{AUC (in \%)} & \tiny \textbf{AP (in \%)} & \tiny \textbf{AUC (in \%)} & \tiny \textbf{AP (in \%)} & \tiny \textbf{AUC (in \%)} & \tiny \textbf{AP (in \%)}\\
    \midrule
    \midrule
    Linear AE (ours) & \underline{90.11} $\pm$ \underline{0.40} & \underline{93.15} $\pm$ \underline{0.28} & \underline{94.64} $\pm$ \underline{0.08} & \underline{95.96} $\pm$ \underline{0.10} & \underline{98.34} $\pm$ \underline{0.03} & \underline{98.46} $\pm$ \underline{0.03} \\
     2-layer GCN AE& 90.29 $\pm$ 0.39 & 93.01 $\pm$ 0.33 & 94.80 $\pm$ 0.08 & 95.72 $\pm$ 0.05 & 97.97 $\pm$ 0.09 & 98.12 $\pm$ 0.09 \\
     3-layer GCN AE& 89.91 $\pm$ 0.61 & 92.24 $\pm$ 0.67 & 94.51 $\pm$ 0.31 & 95.11 $\pm$ 0.28 & 94.35 $\pm$ 1.30 & 94.46 $\pm$ 1.31 \\
    \midrule
    Linear VAE (ours) & \underline{90.62} $\pm$ \underline{0.30} & \underline{93.25} $\pm$ \underline{0.22} & \underline{95.20} $\pm$ \underline{0.16} & \underline{95.99} $\pm$ \underline{0.12} & \underline{98.35} $\pm$ \underline{0.05} & \underline{98.46} $\pm$ \underline{0.05} \\
     2-layer GCN VAE & 90.40 $\pm$ 0.43 & 93.09 $\pm$ 0.35 & 94.60 $\pm$ 0.20 & 95.74 $\pm$ 0.13 & 97.75 $\pm$ 0.08 & 97.91 $\pm$ 0.06 \\
     3-layer GCN VAE & 89.92 $\pm$ 0.59 & 92.52 $\pm$ 0.48 & 94.48 $\pm$ 0.28 & 95.30 $\pm$ 0.22 & 94.57 $\pm$ 1.14 & 94.73 $\pm$ 1.12 \\
    \midrule
    \midrule
    & \multicolumn{2}{c}{\textbf{Blogs}} & \multicolumn{2}{c}{\textbf{Proteins}} & \multicolumn{2}{c}{\textbf{Google}} \\
        \textbf{Model}     & \multicolumn{2}{c}{(n = 1 224, m = 19 025)} & \multicolumn{2}{c}{(n = 6 327, m = 147 547)} & \multicolumn{2}{c}{(n = 15 763, m = 171 206)} \\
     \cmidrule{2-7}
     & \tiny \textbf{AUC (in \%)} & \tiny \textbf{AP (in \%)} & \tiny \textbf{AUC (in \%)} & \tiny \textbf{AP (in \%)} & \tiny \textbf{AUC (in \%)} & \tiny \textbf{AP (in \%)}\\
    \midrule
    \midrule
    Linear AE (ours) & \underline{91.71} $\pm$ \underline{0.39} & \underline{92.53} $\pm$ \underline{0.44} & 94.09 $\pm$ 0.23 & 96.01 $\pm$ 0.16 & 96.02 $\pm$ 0.14 & 97.09 $\pm$ 0.08 \\
     2-layer GCN AE& 91.57 $\pm$ 0.34 & 92.51 $\pm$ 0.29 & 94.55 $\pm$ 0.20 & 96.39 $\pm$ 0.16 & 96.66 $\pm$ 0.24 & 97.45 $\pm$ 0.25 \\
     3-layer GCN AE& 91.74 $\pm$ 0.37 & 92.62 $\pm$ 0.31 & 94.30 $\pm$ 0.19 & 96.08 $\pm$ 0.15 & 95.10 $\pm$ 0.27 & 95.94 $\pm$ 0.20 \\
    \midrule
    Linear VAE (ours) & 91.34 $\pm$ 0.24 & 92.10 $\pm$ 0.24 & 93.99 $\pm$ 0.10 & \underline{95.94} $\pm$ \underline{0.16} & 91.11 $\pm$ 0.31 & 92.91 $\pm$ 0.18 \\
     2-layer GCN VAE& 91.85 $\pm$ 0.22 & 92.60 $\pm$ 0.25 & 94.57 $\pm$ 0.18 & 96.18 $\pm$ 0.33 & 96.11 $\pm$ 0.59 & 96.84 $\pm$ 0.51 \\
     3-layer GCN VAE& 91.83 $\pm$ 0.48 & 92.65 $\pm$ 0.35 & 94.27 $\pm$ 0.25 & 95.71 $\pm$ 0.28 & 95.10 $\pm$ 0.54 & 96.00 $\pm$ 0.44 \\
    \bottomrule
  \end{tabular}
  \end{tiny}

\end{table} 
  \vspace{-0.2cm}
\section{Conclusion}
We highlighted that, in graph AE/VAE, simple first-order linear encoders are as effective as the popular GCNs on numerous real-world graphs. Theses results are consistent with recent efforts, out of the AE/VAE unsupervised frameworks, to simplify GCNs \cite{wu2019simplifying}. Arguing that current benchmark datasets might be too simple, we hope that our analysis will initiate further discussions on the training and evaluation of graph AE/VAE that should, in the future, lead towards their improvement.

\bibliographystyle{ACM-Reference-Format}
\bibliography{sample-base}

\newpage

\section*{Supplementary Material}

This supplementary material provides details on our experiments and two complementary tables.

\subsection*{Annex A - Experimental Setting and Hyperparameters Details}

For \textit{link prediction} experiments (Tables 1, 2 and 3), we followed the setting of \cite{kipf2016-2} and trained models on incomplete versions of graphs where $15\%$ of edges were randomly removed. We created validation and test sets from respectively  $5\%$ and $10\%$ removed edges and from the same number of randomly sampled pairs of unconnected nodes. As \cite{kipf2016-2}, we ignored edges directions when initial graphs were directed. The link prediction task is a binary classification task, consisting in discriminating edges from non-edges in the test set. We note that, due to high performances on the Proteins graph \cite{konect} (AUC and AP scores above 99.5\% for all models), we complicated the learning task, by only including 10\% of edges (resp. 25\%) to the training (resp. test) set for this graph.

All models were trained for $200$ epochs. For Cora, Citeseer and Pubmed, we set identical hyperparameters w.r.t. \cite{kipf2016-2} in order to reproduce their results, i.e. we had $d = 16$, GCNs included 32-dim hidden layers, and we used Adam optimizer with a learning rate of 0.01. We indeed managed to reach similar scores as \cite{kipf2016-2} when training 2-layer GCN-based AE and VAE with, however, larger standard errors. This difference comes from the fact that we report average scores on 100 different and random train/validation/test splits, while they used fixed dataset splits for all runs (randomness in their results comes from initialization). For other datasets, validation set was used for hyperparameters tuning. We adopted a learning rate of 0.1 for Arxiv-HepTh ; of 0.01 for Blogs, Cora-larger, DBLP, Google and Hamsterster and Proteins (AE models) ; of 0.005 for WebKD (except linear AE and VAE where we used 0.001 and 0.01) and Proteins (VAE models). All models learned 16-dim embeddings i.e. $d =16$, with 32-dim hidden layers and without dropout. 

While running time is not the main focus of this paper, we also note that linear AE and VAE models tend to be 10\% to 15\% faster than their GCN-based counterparts, as the proposed simple encoders include fewer matrix operations: e.g. 6.03 seconds (vs 6.73 seconds) mean running time for linear graph VAE (vs 2-layer GCN-based graph VAE) on the featureless Citeseer dataset, on an NVIDIA GTX 1080 GPU. As the standard inner-product decoder has a $O(d n^2)$ quadratic time complexity \cite{salha2019-1}, we focused on medium-size
datasets (with roughly $<30K$ nodes) in our experiments ; we nonetheless point out the existence of recent works \cite{salha2019-1} on scalable graph autoencoders.

\subsection*{Annex B - Experiments on Link Prediction with Alternative Decoders}

In Table 3, we report complementary \textit{link prediction} experiments, on variants of graph AE and VAE where we replaced the inner product decoder by two more complex decoding schemes from existing literature: the Graphite model from \cite{grover2019graphite} and the gravity-inspired asymmetric decoder from \cite{salha2019-2}.

We draw similar conclusions w.r.t. Tables 1 and 2, consolidating results from Section 4. For the sake of brevity, we only report results for the Cora, Citeseer and Pubmed graphs, where linear models are mostly competitive, and for the Google graph, were GCN-based graph AE and VAE are outperforming. We stress out that scores from Graphite and gravity-inspired models are \textit{not} directly comparable, as the former ignores edges directionalities while the latter processes directed graphs (i.e. the learning task becomes a \textit{directed} link prediction problem).

\subsection*{Annex C - Experiments on Node Clustering}

In Table 4, we report \textit{node clustering} experiments on datasets that include node-level ground-truth communities. We trained models on complete graphs, then ran $k$-means algorithms (with $k$-means++ initialization)
in node embedding spaces. We compared output clusters to ground-truth communities by computing \textit{adjusted mutual information (AMI)} scores, averaged over 100 runs.

Overall, linear AE and VAE models are competitive w.r.t. their GCN-based counterparts from \cite{kipf2016-2} on the Cora, Cora-larger, Citeseer and Pubmed citation graphs, in which nodes are documents clustered in respectively 6, 70, 7 and 3 topic classes. However, 2-layer and 3-layer GCN-based models are significantly outperforming on the \textit{dense} Blogs graph, where political blogs are classified as either left-leaning or right-leaning. This is consistent w.r.t. insights from Section 4 that GCN-based encoding should bring larger empirical benefits on dense graphs.

\begin{table}[h]
  \centering
    \caption{Link prediction with alternative decoding schemes \cite{grover2019graphite,salha2019-2}. Performances of linear graph AE/VAE are \underline{underlined} when reaching competitive results  w.r.t. GCN-based models ($\pm$ 1 st. dev.).}
\begin{tiny}
  \begin{tabular}{c|cc|cc|cc|cc}
    \toprule
    & \multicolumn{2}{c}{\textbf{Cora}} & \multicolumn{2}{c}{\textbf{Citeseer}} & \multicolumn{2}{c}{\textbf{Pubmed}} & \multicolumn{2}{c}{\textbf{Google}} \\
     \textbf{Model} & \multicolumn{2}{c}{(n = 2 708, m = 5 429)} & \multicolumn{2}{c}{(n = 3 327, m = 4 732)} & \multicolumn{2}{c}{(n = 19 717, m = 44 338)} & \multicolumn{2}{c}{(n = 15 763, m = 171 206)} \\
     \cmidrule{2-9}
      & \tiny \textbf{AUC (in \%)} & \tiny \textbf{AP (in \%)} & \tiny \textbf{AUC (in \%)} & \tiny \textbf{AP (in \%)} & \tiny \textbf{AUC (in \%)} & \tiny \textbf{AP (in \%)} & \tiny \textbf{AUC (in \%)} & \tiny \textbf{AP (in \%)}\\
    \midrule
    \midrule
    Linear Graphite AE (ours) & \underline{83.42} $\pm$ \underline{1.76} & \underline{87.32} $\pm$ \underline{1.53} & \underline{77.56} $\pm$ \underline{1.41} & \underline{82.88} $\pm$ \underline{1.15} & \underline{80.28} $\pm$ \underline{0.86} & \underline{85.81} $\pm$ \underline{0.67} & 94.30 $\pm$ 0.22  & 95.09 $\pm$ 0.16 \\
     2-layer Graphite AE & 81.20 $\pm$ 2.21 & 85.11 $\pm$ 1.91 & 73.80 $\pm$ 2.24 & 79.32 $\pm$ 1.83 & 79.98 $\pm$ 0.66 & 85.33 $\pm$ 0.41 & 95.54 $\pm$ 0.42  & 95.99 $\pm$ 0.39 \\
     3-layer Graphite AE & 79.06 $\pm$ 1.70 & 81.79 $\pm$ 1.62 & 72.24 $\pm$ 2.29 & 76.60 $\pm$ 1.95 & 79.96 $\pm$ 1.40 & 84.88 $\pm$ 0.89 & 93.99 $\pm$ 0.54  & 94.74 $\pm$ 0.49 \\
    \midrule
    Linear Graphite VAE (ours) & \underline{83.68} $\pm$ \underline{1.42} & \underline{87.57} $\pm$ \underline{1.16} & \underline{78.90} $\pm$ \underline{1.08} & \underline{83.51} $\pm$ \underline{0.89} & 79.59 $\pm$ 0.33 & 86.17 $\pm$ 0.31 & 92.71 $\pm$ 0.38  & 94.41 $\pm$ 0.25 \\
     2-layer Graphite VAE  & 84.89 $\pm$ 1.48 & 88.10 $\pm$ 1.22 & 77.92 $\pm$ 1.57 & 82.56 $\pm$ 1.31 & 82.74 $\pm$ 0.30 & 87.19 $\pm$ 0.36 & 96.49 $\pm$ 0.22  & 96.91 $\pm$ 0.17 \\
     3-layer Graphite VAE  & 85.33 $\pm$ 1.19 & 87.98 $\pm$ 1.09 & 77.46 $\pm$ 2.34 & 81.95 $\pm$ 1.71 & 84.56 $\pm$ 0.42 & 88.01 $\pm$ 0.39 & 96.32 $\pm$ 0.24  & 96.62 $\pm$ 0.20 \\
    \midrule
    \midrule
    Linear Gravity AE (ours) & \underline{90.71} $\pm$ \underline{0.95} & \underline{92.95} $\pm$ \underline{0.88} & \underline{80.52} $\pm$ \underline{1.37} & \underline{86.29} $\pm$ \underline{1.03} & \underline{76.78} $\pm$ \underline{0.38} & \underline{84.50} $\pm$ \underline{0.32} & 97.46 $\pm$ 0.07  & 98.30 $\pm$ 0.04 \\
     2-layer Gravity AE & 87.79 $\pm$ 1.07 & 90.78 $\pm$ 0.82 & 78.36 $\pm$ 1.55 & 84.75 $\pm$ 1.10 & 75.84 $\pm$ 0.42 & 83.03 $\pm$ 0.22 & 97.77 $\pm$ 0.10  & 98.43 $\pm$ 0.10 \\
     3-layer Gravity AE & 87.76 $\pm$ 1.32 & 90.15 $\pm$ 1.45 & 78.32 $\pm$ 1.92 & 84.88 $\pm$ 1.36 & 74.61 $\pm$ 0.30 & 81.68 $\pm$ 0.26 & 97.58 $\pm$ 0.12  & 98.28 $\pm$ 0.11 \\
    \midrule
    Linear Gravity VAE (ours) & \underline{91.29} $\pm$ \underline{0.70} & \underline{93.01} $\pm$ \underline{0.57} & \underline{86.65} $\pm$ \underline{0.95} & \underline{89.49} $\pm$ \underline{0.69} & \underline{79.68} $\pm$ \underline{0.36} & \underline{85.00} $\pm$ \underline{0.21} & 97.32 $\pm$ 0.06  & \underline{98.26} $\pm$ \underline{0.05} \\
     2-layer Gravity VAE  & 91.92 $\pm$ 0.75 & 92.46 $\pm$ 0.64 & 87.67 $\pm$ 1.07 & 89.79 $\pm$ 1.01 & 77.30 $\pm$ 0.81 & 82.64 $\pm$ 0.27 & 97.84 $\pm$ 0.25  & 98.18 $\pm$ 0.14 \\
     3-layer Gravity VAE  & 90.80 $\pm$ 1.28 & 92.01 $\pm$ 1.19 & 85.28 $\pm$ 1.33 & 87.54 $\pm$ 1.21 & 76.52 $\pm$ 0.61 & 80.73 $\pm$ 0.63 & 97.32 $\pm$ 0.23  & 97.81 $\pm$ 0.20 \\
    \bottomrule
  \end{tabular}
  \end{tiny}
\end{table}

\begin{table}[h]
  \centering
\begin{tiny}
    \caption{Node Clustering on graphs with communities. Performances of linear graph AE/VAE are \underline{underlined} when reaching competitive results w.r.t. GCN-based models \cite{kipf2016-2} ($\pm$ 1 st. dev.).}
  \begin{tabular}{c|c|c|c|c}
    \toprule
    & \textbf{Cora} & \textbf{Citeseer} & \textbf{Pubmed} & \textbf{Cora-larger}\\
      \textbf{Model}  & \tiny (n = 2.708, m = 5 429) & \tiny (n = 3 327, m = 4 732) & \tiny (n = 19 717, m = 44 338) & \tiny (n = 23 166, m = 91 500)\\
     \cmidrule{2-5}
      & \tiny \textbf{AMI (in \%)} & \tiny \textbf{AMI (in \%)} & \tiny \textbf{AMI (in \%)} & \tiny \textbf{AMI (in \%)} \\
    \midrule
    \midrule
    Linear AE (ours) & 26.31 $\pm$ 2.85 & \underline{8.56} $\pm$ \underline{1.28} & 10.76 $\pm$ 3.70 & \underline{40.34} $\pm$ \underline{0.51} \\
     2-layer GCN AE & 30.88 $\pm$ 2.56 & 9.46 $\pm$ 1.06 & 16.41 $\pm$ 3.15 & 39.75 $\pm$ 0.79 \\
     3-layer GCN AE & 33.06 $\pm$ 3.10 & 10.69 $\pm$ 1.98 & 23.11 $\pm$ 2.58 & 35.67 $\pm$ 1.76 \\
     \midrule
         Linear VAE (ours) &  \underline{34.35} $\pm$ \underline{1.42} & \underline{12.67} $\pm$ \underline{1.27} & \underline{25.14} $\pm$ \underline{2.83} & \underline{43.32} $\pm$ \underline{0.52} \\
     2-layer GCN VAE & 26.66 $\pm$ 3.94 & 9.85 $\pm$ 1.24 & 20.52 $\pm$ 2.97 & 38.34 $\pm$ 0.64 \\
     3-layer GCN VAE &  28.43 $\pm$ 2.83 & 10.64 $\pm$ 1.47 & 21.32 $\pm$ 3.70 & 37.30 $\pm$ 1.07 \\
    \midrule
    \midrule
        & \textbf{Cora, with features} & \textbf{Citeseer, with features} & \textbf{Pubmed, with features} & \textbf{Blogs}\\
        \textbf{Model} & \tiny (n = 2 708, m = 5 429, f = 1 433) & \tiny (n = 3 327, m = 4 732, f = 3 703) & \tiny (n = 19 717, m = 44 338, f = 500) & \tiny (n = 1 224, m = 19 025)\\
     \cmidrule{2-5}
      & \tiny \textbf{AMI (in \%)} & \tiny \textbf{AMI (in \%)} & \tiny \textbf{AMI (in \%)} & \tiny \textbf{AMI (in \%)}\\
    \midrule
    \midrule
    Linear AE (ours) & \underline{47.02} $\pm$ \underline{2.09} & \underline{20.23} $\pm$ \underline{1.36} & \underline{26.12} $\pm$ \underline{1.94} & 46.84 $\pm$ 1.79 \\
     2-layer GCN AE & 43.04 $\pm$ 3.28 & 19.38 $\pm$ 3.15 & 23.08 $\pm$ 3.35 & 72.58 $\pm$ 4.54  \\
     3-layer GCN AE & 44.12 $\pm$ 2.48 & 19.71 $\pm$ 2.55 & 25.94 $\pm$ 3.09 & 72.72 $\pm$ 1.80 \\
\midrule
    Linear VAE (ours) & \underline{48.12} $\pm$ \underline{1.96} & \underline{20.71} $\pm$ \underline{1.95} & \underline{29.74} $\pm$ \underline{0.64} & 49.70 $\pm$ 1.08 \\
     2-layer GCN VAE& 44.84 $\pm$ 2.63 & 20.17 $\pm$ 3.07 & 25.43 $\pm$ 1.47 & 73.12 $\pm$ 0.83 \\
     3-layer GCN VAE & 44.29 $\pm$ 2.54 & 19.94 $\pm$ 2.50 & 24.91 $\pm$ 3.09 & 70.56 $\pm$ 5.43 \\
    \bottomrule
  \end{tabular}
  \end{tiny}
\end{table} 
\end{document}